\documentclass[runningheads]{llncs}
\usepackage[T1]{fontenc}
\usepackage{graphicx}
\usepackage{color}
\usepackage{cite}
\usepackage{amsmath,amssymb,amsfonts}

\DeclareMathOperator*{\argmin}{arg\,min}

\begin{document}

\title{Unsupervised Echocardiography Registration through Patch-based MLPs and Transformers}
\titlerunning{Patch-based registration}
\author{Zihao Wang$^*$, Yingyu Yang$^*$, Maxime Sermesant, Herv\'e Delingette}
\authorrunning{Z. Wang, Y. Yang et al.}
\institute{Inria Sophia Antipolis, Epione Lab., Université Côte d’Azur, Valbonne, France\\
\email{Contact: \{firstname\}.\{lastname\}@inria.fr}}
\maketitle              
\def\thefootnote{*}\footnotetext{These authors contributed equally to this work}\def\thefootnote{\arabic{footnote}}

\begin{abstract}
Image registration is an essential but challenging task in medical image computing, especially for echocardiography, where the anatomical structures are relatively noisy compared to other imaging modalities. Traditional (non-learning) registration approaches rely on the iterative optimization of a similarity metric which is usually costly in time complexity. In recent years, convolutional neural network (CNN) based image registration methods have shown good effectiveness. In the meantime, recent studies show that the attention-based model (e.g., Transformer) can bring superior performance in pattern recognition tasks. In contrast, whether the superior performance of the Transformer comes from the long-winded architecture or is attributed to the use of patches for dividing the inputs is unclear yet. This work introduces three patch-based frameworks for image registration using MLPs and transformers. We provide experiments on 2D-echocardiography registration to answer the former question partially and provide a benchmark solution. Our results on a large public 2D echocardiography dataset show that the patch-based MLP/Transformer model can be effectively used for unsupervised echocardiography registration. They demonstrate comparable and even better registration performance than a popular CNN registration model. In particular, patch-based models better preserve volume changes in terms of Jacobian determinants, thus generating robust registration fields with less unrealistic deformation. Our results demonstrate that patch-based learning methods, whether with attention or not, can perform high-performance unsupervised registration tasks with adequate time and space complexity. Our codes are available \footnote{\url{https://gitlab.inria.fr/epione/mlp\_transformer\_registration}}.

\keywords{Unsupervised Registration \and MLP \and Transformer \and Echocardiography.}
\end{abstract}
%
%
%=========================================================================================================
\section{Introduction}
Image registration is essential for clinical usage; for example, the registration of cardiac images between end-diastole and end-systole is meaningful in myocardium deformation analysis. 
Non-rigid echocardiography image registration is one of the most challenging image registration tasks, as finding the deformation field between noisy images is a highly nonlinear problem in the absence of ground truth deformation. Specifically, various image registration problems require the mapping between moving and fixed images to be folding-free~\cite {diffeo1,diffeo2,diffeo3}. Traditional non-learning approaches rely on optimizing similarity metrics to measure the matching quality between image pairs \cite{Intro_Reg_review,DAVATZIKOS1997207,XavierPennecDemons}.
With the rapid promotion of deep learning, various frameworks of convolutional neural networks (CNN) have been introduced in image registration and have shown impressive performance in many research works.

We consider a 2D non-rigid machine learning-based  image registration task in this work. With two given images: $I_{fix}^{N}, N \in \mathbb{R}$ and $I_{mov}^{N}, N \in \mathbb{R}$, we want to learn a model $ \mathcal{T}_{\omega}(I_{mov}, I_{fix}) \rightarrow \phi(\theta)$ that generates a constrained transformation $\phi(\theta)$ based on a similarity measurement $\mathcal{M}$ to warp the moving image by minimizing the loss function:
\begin{equation}
L = \argmin_{\theta} \mathcal{M}(I_{fix}, I_{mov} \circ \phi(\theta)) + \lambda\mathcal{C}(\phi(\theta))
\label{eq:loss}
\end{equation}
where the transformation $\phi$ is parameterised by the parameter $\theta$ and constrained by a regularisation term ${C}(\phi(\theta))$ to ensure $\phi$ to be a spatially smooth transformation.
However, iterative optimization of Eq. \ref{eq:loss} is very time-consuming, whereas a well-trained CNN does not need any iterative minimization of the loss function at test time. This advantage drives researchers' attention to learning-based registration. Learning-based registration methods can be categorized into supervised and unsupervised registration approaches.

\paragraph{Supervised Registration}
The supervised learning registration methods \cite{Intro_DL_Wu,Intro_DL_CAO,SVFNet} are primarily trained on a ground-truth training set for which simulated or registered displacement fields are available. The training dataset is usually generated with traditional registration frameworks or by generating artificial deformation fields as ground truth for warping the moving images to get the fixed images. \cite{SokootiIntro,YANG2017378}. 
One of the limitations of the supervised registration approaches is the registration quality, which is highly influenced by the nature of the training set of the deformation map, although the requirement in terms of the training set can be partially alleviated by using weakly-supervised learning \cite{HU20181,BLENDOWSKI2021101822,voxelmorph,8363756,8458414}.

\paragraph{Unsupervised Registration}
In unsupervised registration \cite{Intro_DL_Krebs,DALCA2019226,voxelmorph,HERING2021102139,MANSILLA2020269}, we rely on a similarity measure and regularisation to optimize the neural network for learning the transformations between the fixed and moving images. 
Usually, a CNN is used directly for warping the moving images, which is then compared to the fixed image with the similarity loss. 
The displacement field can also be obtained from a generative adversarial neural network, which introduces a discriminator neural network for assessing the generated deformation field quality. \cite{GAN1,GAN2,GAN3,GAN4}.

\paragraph{Multi-layer Perceptron and Transformers}
MLP is one of the most classical neural networks and consists of a stack of linear layers along with non-linear activation \cite{mlp1}. 
For several years, CNN has been widely used due to its performance on vision tasks and its computation efficiency \cite{cnn1}. Recently, several alternatives to the CNN have been proposed such as  Vision Transformer (ViT) \cite{Dosovitskiy2021AnII} or MLP-Mixer \cite{mlpmixer} which  demonstrated comparable or even better performance than CNN on classification or detection tasks. There are currently intense discussions in the community of whether patching, attention or simple MLP plays the most important role in a such good performance. 

In this paper, we propose three MLP/Transformer based models for echocardiography image registration and compare them with one representative CNN model in unsupervised echocardiography image registration. There are already works using transformers to register medical images, such as TransMorph~\cite{chen2021transmorph} and Dual Transformer~\cite{zhang2021learning}, but they are mostly restricted to high signal-to-noise medical images, such as MRI images and CT images. While ultrasound images (2D) are actually the most popular imaging modality in the real world. Our inspiration not only comes from the trending debate over Transformer and MLP but also stems from the intuition that patch-based learning methods share similar logic to a traditional block-matching method for cardiac tracking. 

Our contributions are two-folded. First, we show the effectiveness of patch-based MLP/Transformer models in medical image registration compared with a CNN-based registration model. Second, we conduct a thorough ablation study of the influence of different structures (MLP, MLP-Mixer, Transformer) and different scales (single scale or multiple scales). Our results provide empirical support to the observation that the attention mechanism may not be the only key factor in the SOTA performances. \cite{liu2021pay,melaskyriazi2021need}, at least in the field of unsupervised image registration.

%==========================================================================================================
\section{Methodology}
\label{method}

\subsection{Diffeomorphic Registration}
We estimate a diffeomorphic transform between images, which preserves topology and is folding-free. Our model generates stationary velocity field $v(\theta)$ \cite{ASHBURNER2000805} instead of generating displacement maps, thanks to an integration layer applied to the velocity field leading to diffeomorphism $\phi(\theta)$. Formally, the diffeomorphic transformation $\phi$ is the solution to a differential equation related to the predicted (stationary) velocity field  $V$~\cite{DALCA2019226}:$\frac{\partial \phi_t}{\partial t} = v(\phi_t); \phi_{t=0} = Id$. 
In a stationary velocity field setting, the transformation $\phi$ is defined as the exponential of the velocity field $\phi = \exp (v)$ \cite{diffeo3}. The integration (exponential) layer applies the scaling and squaring method to approximate the diffeomorphic transform \cite{julian18}. The obtained transformation $\phi$ is then used by a spatial transform layer to deform the image.

\subsection{Proposed frameworks}
\label{sec:models}
\begin{figure}[!ht]
    \centering
    \includegraphics[width=1\linewidth]{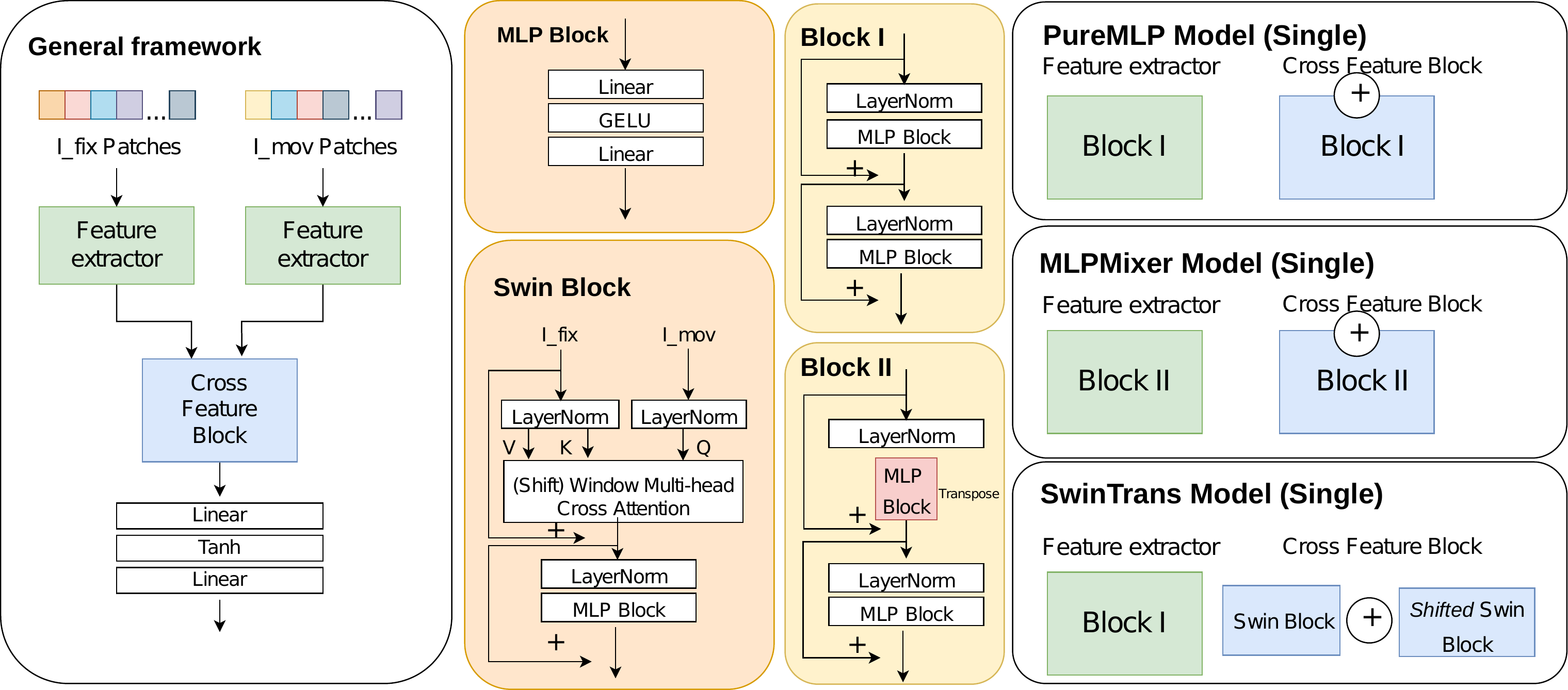}
    \caption{The detailed composition of proposed three frameworks. Here we only show single-scale models. Please read Section.\ref{sec:models} for more description.}
    \vspace{-2mm}
    \label{fig:models}
\end{figure}

Given two images, $I_{fix}$ and $I_{mov}$, we would to estimate the transformation $\phi(\theta)$ that transforms the moving image to the fixed image so that $I_{fix} \approx I_{mov} \circ \phi(\theta)$. We approximate the ideal $\phi(\theta)$ by the following proposed frameworks. The following three propositions are all based on patch-wise manipulations and share a similar general architecture. As shown in Fig.\ref{fig:models}, $I_{fix}$ and $I_{mov}$ are both processed by an identical feature extractor (green block) separately. The two feature maps are then passed through the cross-feature block (blue block). After two linear layers, we obtain their corresponding velocity field. The velocity field passes through an integration layer and we obtain the final displacement field by up-sampling it to the original image size. 

\paragraph{\textbf{Pure MLP registration framework}}
The same MLP block (Block I in Fig.\ref{fig:models}) is used for the feature extractor and the cross-feature block in this model. The outputs from two separate feature extractors (shared weights) are added together before feeding into the cross-feature block. 
We note this model \textbf{PureMLP} for abbreviation in the following paper. 

\paragraph{\textbf{MLP-Mixer registration framework}}
The MLP-Mixer registration framework is very similar to the former Pure MLP framework. The only difference is that the three MLP blocks used for separate feature extraction and cross-feature processing are replaced by MLP-Mixer blocks~\cite{mlpmixer} (Block II in Fig.\ref{fig:models}). The MLP-Mixer block has an identical structure to the MLP block, but with a feature map transpose to obtain channel-wise feature fusion (the red cell of Block II in Fig.\ref{fig:models}). We note this model \textbf{MLPMixer} for abbreviation in the following paper. 

\paragraph{\textbf{Swin-Transformer registration framework}}
This model uses the MLP block (Block I in Fig.\ref{fig:models}) to first extract patch-based features for both $I_{fix}$ and $I_{mov}$. For cross feature block, we adapt the recent Swin Transformer \cite{liu2021Swin} to do the cross-patch attention locally (Swin Block in Fig.\ref{fig:models}). Our Swin block accepts feature input from both images ($I_{fix}$ and $I_{mov}$), where key $K$ and value $V$ are normalized $I_{fix}$ features while query $Q$ comes from normalized $I_{mov}$ features. Swin block calculates the cross-attention within a pre-defined window region. We perform normal window partition for $I_{fix}$ features and one normal partition, one shifted partition for $I_{mov}$ features. The cross-attention under the two types of window partition configurations is summed together before feeding to the final linear layers. Due to the page limit, we invite interested readers to refer to~\cite{liu2021Swin} for a detailed description of the Swin transformer mechanism. We note this model \textbf{SwinTrans} for abbreviation in the following paper. 

\subsection{Multi-scale features}
In order to enforce different reception fields for patch-based models, we decide to combine multi-scale models together. This is accomplished by adopting models of different patch sizes together. It is quite similar to how CNN achieves this goal, by applying a larger kernel or adding pooling layers. In particular, our multi-scale model consists of several parallel independent single-scale child models. The output of each child model is upsampled and then combined together to form the final estimation of velocity field $v(\theta)$ 
\begin{equation}
    v(\theta) = \sum_{C=1}^{N} \omega_C Out_C
    \label{sum_weight}
\end{equation}
where $Out_C$ is the output of child model $C$. The final velocity field $v$ is then passed to calculate the final transform $\phi$ as depicted in former subsections.

%====================================================================================================
\section{Experiments and Results}
\label{experiment}
\subsection{Dataset}
To evaluate the effectiveness of our unsupervised registration models, we use a publicly accessible 2D echocardiography dataset CAMUS\footnote{https://www.creatis.insa-lyon.fr/Challenge/camus/}. This dataset consists of 500 patients, each having 2D apical 4-chamber (A4C) and 2-chamber (A2C) view sequences. Manual annotation of cardiac structures (left endocardium, left epicardium and left atrium) was acquired by expert cardiologists for each patient in each view, at end-diastole (ED) and end-systole (ES)~\cite{Leclerc2019b}. The structure annotations of 450 patients are publicly available while that of the other 50 patients are unreleased. In total, we have 1000 pairs of ED/ES images and we randomly split (still considering age and image quality distribution) the 900 pairs (with annotations) into training (630), validation(90), and test data (180). The 100 pairs (without annotations) are included into the training set (730).  

\subsection{Implementation}
We compare our proposed three models with a very popular CNN registration model VoxelMorph \cite{voxelmorph}. To be consistent with our setting, we make use of the diffeomorphic version of the VoxelMorph model (we use the abbreviation \textbf{Vxm} in the following paper). We train all the models with input images resized to 128x128 pixels and use an Adam optimizer (learning rate = 0.0001). We set the training epoch to be 500 and training is early stopped when there is no improvement on the validation set over 30 epochs. 

\paragraph{Loss function}
In order to enforce the diffeomorphic property of our registration model, we apply a symmetric loss function for all the unsupervised models:
\begin{equation}
    \argmin L = L_{mse}(\hat\phi(I_{move}), I_{fix}) +  L_{mse}(\hat\phi^{-1}(I_{fix}), I_{move}) + \lambda * L_{diff} (\hat\phi)
\end{equation}

where $\hat\phi^{-1}$ is the inverse of $\hat\phi$ and $L_{diff}$ is a diffusion regularizer for smoothness $L_{diff} = \int ||\nabla_x \phi + \nabla_y \phi||^2$ and set $\lambda = 0.01$ according to  \cite{voxelmorph}.   

\paragraph{Data augmentation}
In order to improve model generalization and avoid outfitting, we apply the same random data augmentation tricks for each image pair during the training phase. The following augmentation techniques: rotation, cropping and resizing, brightness adjustment, contrast change, sharpening, blurring, and speckle noise addition are conducted with a probability of 0.5 separately. No augmentation is applied during the validation or test phase. 

\subsection{Experiments}
\paragraph{\textbf{Multi-scale models}} (abbreviation: model name + \_M) we apply three child models for PureMLP, MLPMixer and SwinTrans (with patch sizes of 4x4, 8x8, and 16x16 respectively). For child models of size 4x4, 8x8, and 16x16 in SwinTrans, we set the number of window sizes to 8, 4, and 2, and the number of heads to 32, 16, and 8 respectively. The dimension of patch embedding is set to be 128 for all patch-based methods. $\omega_C$ in Equation.\ref{sum_weight} is set to be 0.5, 0.3, 0.2 for the child model with patch sizes of 4x4, 8x8, and 16x16 separately.
\paragraph{\textbf{Single-scale models}} (abbreviation: model name + \_S) we run single-scale models for PureMLP, MLPMixer, and SwinTrans three proposed frameworks (using patch size of 4x4 pixels). The same configuration is set as for the child model with patch-size 4x4 in multi-scale models. 

\subsection{Results}
Since our SwinTrans model relies mostly on features of $I_{fix}$ (with skip-connection of $I_{fix}$ features), we report only the metrics related to the transformation $\phi(\theta)$ that $I_{fix} \approx I_{move} \circ \phi(\theta)$. For the CAMUS test dataset, we report the Dice score, Hausdorff distance (HD), and mean surface distance (MSD) between the ground truth ED mask and transformed ES mask and the Jacobian determinant in the area of myocardium region. 

\begin{figure}[!ht]
    \centering
    \includegraphics[width=1\linewidth]{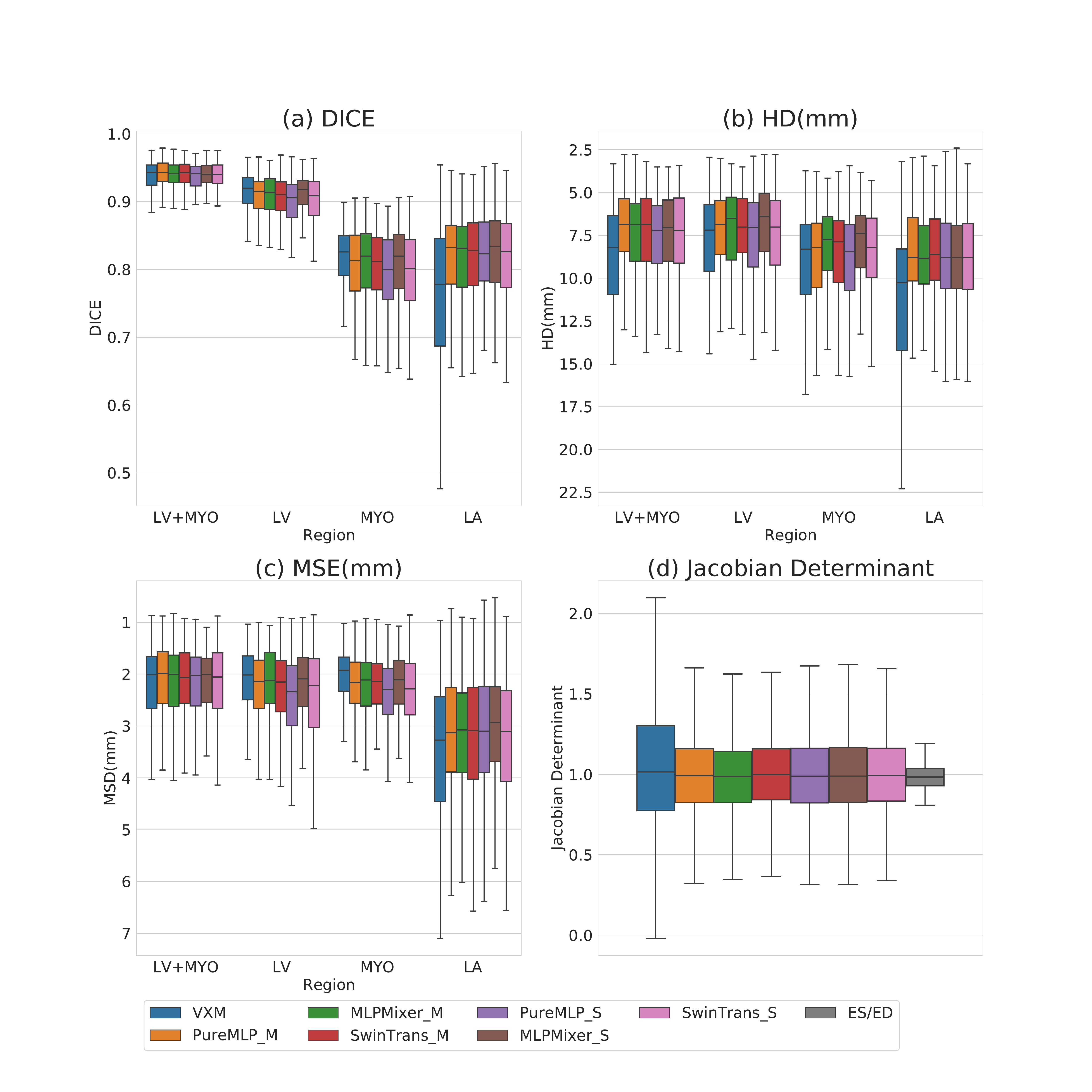}
    \caption{Comparison of evaluation metrics (Dice score, Hausdorff distance (HD), mean surface distance (MSD), and Jacobin determinant) on a test dataset of CAMUS. The Jacobin determinant is only computed in the myocardium region. Except for the Jacobin determinant figure, the higher the boxplot is in the figure, the better performance it will be. }
    \vspace{-2mm}
    \label{fig:camus_test}
\end{figure}

\begin{figure}[!ht]
    \centering
    \includegraphics[width=1\linewidth]{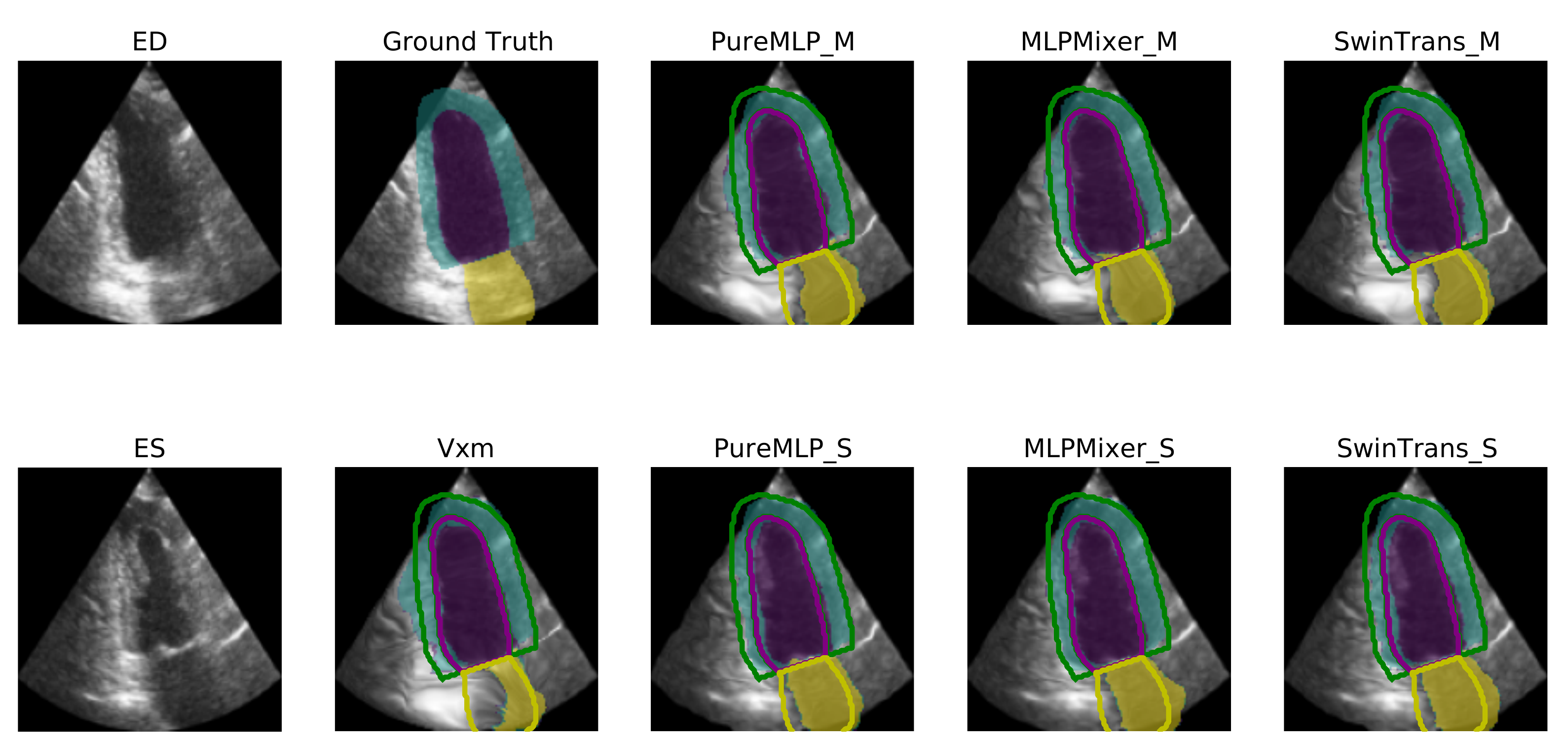}
    \caption{The same registration example on CAMUS test data with transformed ES masks. Colorful patches are corresponding estimations while bold contours are the ground truth (Yellow: left atrium, Purple: left ventricle, Green: myocardium). }
    \label{fig:camus_exp}
\end{figure}

\subsubsection{Evaluation on CAMUS dataset}
From Fig.\ref{fig:camus_test} we can observe that on the CAMUS test dataset, almost all the proposed models, no matter whether it is multi-scale or single-scale, no matter what kind of sub-block it contains (MLP or Transformer or MLP-Mixer), have achieved comparable performance than the CNN model (Vxm), in particular for the whole left ventricle and left atrium registration. In addition, the distribution of the Jacobin determinant shows that our patch-based methods tend to generate a more plausible transform, i.e. closer to real ES/ED myocardium area change. This is consistent with the Hausdorff distance results, which indicates that while preserving comparable registration performance, patch-based methods are more resistant to the estimation of false large deformation (see the atrium and myocardium region of example in Fig.\ref{fig:camus_exp}). What's more, single-scale models and multi-scale models have similar performance. With single-sized patches, we are already capable to let feature information flow through the whole image area and estimating registration transform efficiently (see time and space complexity in Table.\ref{tab:time_space}). 

\begin{table}
\centering
\caption{Time and space complexity  between different models (evaluated on a  GTX 2080Ti)}
\label{tab:time_space}
\begin{tabular}{cccc} 
\hline
Model        & GPU Memory & Train time (s/pair) & Test time (s/pair)  \\ 
\hline
Vxm          & 1365 MiB   & 0.020               & 0.0047              \\
PureMLP\_S   & 1411 MiB   & 0.020               & 0.0038              \\
MLPMixer\_S  & 1447 MiB   & 0.020               & 0.0038              \\
SwinTrans\_S & 1479 MiB   & 0.029               & 0.0047              \\
\hline
\end{tabular}
\end{table}

\section{Conclusion}
In summary, we propose three novel patch-based registration architectures using only MLPs and Transformers. We show that our single and multi-scale models perform similarly and even better to CNN-based registration frameworks on a large echocardiography dataset. The three proposed models demonstrate similar performance among themselves. Our experiments show that patch-based models using MLP/Transformer can perform 2D medical image registration. We shared a similar conclusion with previous works \cite{liu2021pay,melaskyriazi2021need} that the success of Transformer in vision tasks cannot be simply attributed to the attention mechanism, at least in image registration tasks. 
Future works will concentrate on the application of MLP/Transformer in time-series motion tracking.

\subsubsection{Acknowledgements} 
This work has been supported by the French government through the  National Research Agency (ANR)  Investments in the Future with 3IA Côte d’Azur (ANR-19-P3IA-0002) and by Inria PhD funding. 
\bibliographystyle{splncs04}
\bibliography{mybib}

\begin{thebibliography}{10}
\providecommand{\url}[1]{\texttt{#1}}
\providecommand{\urlprefix}{URL }
\providecommand{\doi}[1]{https://doi.org/#1}

\bibitem{Dosovitskiy2021AnII}
Alexey~Dosovitskiy, e.a.: An image is worth 16x16 words: Transformers for image
  recognition at scale. ArXiv  \textbf{abs/2010.11929} (2021)

\bibitem{diffeo3}
Arsigny, V., et~al.: A log-euclidean framework for statistics on
  diffeomorphisms. In: Medical Image Computing and Computer-Assisted
  Intervention -- MICCAI 2006. pp. 924--931. Springer Berlin Heidelberg,
  Berlin, Heidelberg (2006)

\bibitem{ASHBURNER2000805}
Ashburner, J., Friston, K.J.: Voxel-based morphometry—the methods. NeuroImage
   \textbf{11}(6),  805--821 (2000)

\bibitem{voxelmorph}
Balakrishnan, G., et~al.: Voxelmorph: A learning framework for deformable
  medical image registration. IEEE Transactions on Medical Imaging
  \textbf{38}(8),  1788--1800 (2019)

\bibitem{BLENDOWSKI2021101822}
Blendowski, M., et~al.: Weakly-supervised learning of multi-modal features for
  regularised iterative descent in 3d image registration. Medical Image
  Analysis  \textbf{67},  101822 (2021)

\bibitem{Intro_DL_CAO}
Cao, X., et~al.: Deformable image registration based on similarity-steered cnn
  regression. In: Medical Image Computing and Computer Assisted Intervention.
  pp. 300--308. Springer International Publishing, Cham (2017)

\bibitem{diffeo1}
Cao, Y., et~al.: Large deformation diffeomorphic metric mapping of vector
  fields. IEEE Transactions on Medical Imaging  \textbf{24}(9),  1216--1230
  (2005)

\bibitem{chen2021transmorph}
Chen, J., Frey, E.C., He, Y., Segars, W.P., Li, Y., Du, Y.: Transmorph:
  Transformer for unsupervised medical image registration. arXiv preprint
  arXiv:2111.10480  (2021)

\bibitem{DALCA2019226}
Dalca, A.V., et~al.: Unsupervised learning of probabilistic diffeomorphic
  registration for images and surfaces. Medical Image Analysis  \textbf{57},
  226--236 (2019)

\bibitem{DAVATZIKOS1997207}
Davatzikos, C.: Spatial transformation and registration of brain images using
  elastically deformable models. Computer Vision and Image Understanding
  \textbf{66}(2),  207--222 (1997)

\bibitem{GAN4}
Debayle, J., Presles, B.: Rigid image registration by general adaptive
  neighborhood matching. Pattern Recognition  \textbf{55},  45--57 (2016)

\bibitem{8458414}
Ferrante, E., et~al.: Weakly supervised learning of metric aggregations for
  deformable image registration. IEEE Journal of Biomedical and Health
  Informatics  \textbf{23}(4),  1374--1384 (2019)

\bibitem{HERING2021102139}
Hering, A., et~al.: Cnn-based lung ct registration with multiple anatomical
  constraints. Medical Image Analysis  \textbf{72},  102139 (2021)

\bibitem{8363756}
Hu, Y., et~al.: Label-driven weakly-supervised learning for multimodal
  deformable image registration. In: 2018 IEEE 15th International Symposium on
  Biomedical Imaging (ISBI 2018). pp. 1070--1074 (2018)

\bibitem{HU20181}
Hu, Y., et~al.: Weakly-supervised convolutional neural networks for multimodal
  image registration. Medical Image Analysis  \textbf{49},  1--13 (2018)

\bibitem{mlpmixer}
Ilya, et~al.: Mlp-mixer: An all-mlp architecture for vision. CoRR
  \textbf{abs/2105.01601} (2021)

\bibitem{Intro_DL_Krebs}
{Krebs}, J., et~al.: Learning a probabilistic model for diffeomorphic
  registration. IEEE Transactions on Medical Imaging  \textbf{38}(9),
  2165--2176 (2019)

\bibitem{julian18}
Krebs, J., et~al.: {Unsupervised Probabilistic Deformation Modeling for Robust
  Diffeomorphic Registration}. In: Deep Learning in Medical Image Analysis and
  Multimodal Learning for Clinical Decision Support. pp. 101--109. Springer
  International Publishing, Cham (2018)

\bibitem{cnn1}
Krizhevsky, A., et~al.: Imagenet classification with deep convolutional neural
  networks. In: Pereira, F., Burges, C.J.C., Bottou, L., Weinberger, K.Q.
  (eds.) Advances in Neural Information Processing Systems. vol.~25. Curran
  Associates, Inc. (2012)

\bibitem{Leclerc2019b}
{Leclerc}, S., {Smistad}, E., et~al.: Deep learning for segmentation using an
  open large-scale dataset in 2d echocardiography. IEEE Transactions on Medical
  Imaging  \textbf{38}(9),  2198--2210 (2019)

\bibitem{liu2021pay}
Liu, H., et~al.: Pay attention to mlps (2021)

\bibitem{liu2021Swin}
Liu, Z., et~al.: Swin transformer: Hierarchical vision transformer using
  shifted windows. International Conference on Computer Vision (ICCV)  (2021)

\bibitem{GAN3}
Mahapatra, D., Ge, Z.: Training data independent image registration using
  generative adversarial networks and domain adaptation. Pattern Recognition
  \textbf{100},  107109 (2020)

\bibitem{MANSILLA2020269}
Mansilla, L., et~al.: Learning deformable registration of medical images with
  anatomical constraints. Neural Networks  \textbf{124},  269--279 (2020)

\bibitem{melaskyriazi2021need}
Melas-Kyriazi, L.: Do you even need attention? a stack of feed-forward layers
  does surprisingly well on imagenet (2021)

\bibitem{Intro_Reg_review}
Oliveira, F.P., Tavares, J.M.R.: Medical image registration: a review. Computer
  Methods in Biomechanics and Biomedical Engineering  \textbf{17}(2),  73--93
  (2014)

\bibitem{SVFNet}
Roh{\'e}, M.M., et~al.: {SVF-Net: Learning Deformable Image Registration Using
  Shape Matching}. In: {MICCAI 2017 - the 20th International Conference on
  Medical Image Computing and Computer Assisted Intervention}. pp. 266--274.
  Medical Image Computing and Computer Assisted Intervention -- MICCAI 2017,
  {Springer International Publishing}, Qu{\'e}bec, Canada (Sep 2017),
  \url{https://hal.inria.fr/hal-01557417}

\bibitem{SokootiIntro}
Sokooti, H., et~al.: Nonrigid image registration using multi-scale 3d
  convolutional neural networks. In: Medical Image Computing and Computer
  Assisted Intervention - MICCAI 2017. pp. 232--239. Springer (2017)

\bibitem{GAN1}
Tanner, C., et~al.: Generative adversarial networks for mr-ct deformable image
  registration (2018)

\bibitem{mlp1}
Van Der~Malsburg, C.: Frank rosenblatt: Principles of neurodynamics:
  Perceptrons and the theory of brain mechanisms. In: Palm, G., Aertsen, A.
  (eds.) Brain Theory. pp. 245--248. Springer Berlin Heidelberg, Berlin,
  Heidelberg (1986)

\bibitem{XavierPennecDemons}
Vercauteren, T., et~al.: Non-parametric diffeomorphic image registration with
  the demons algorithm. In: Medical Image Computing and Computer-Assisted
  Intervention -- MICCAI 2007. pp. 319--326. Springer Berlin Heidelberg,
  Berlin, Heidelberg (2007)

\bibitem{diffeo2}
Vercauteren, T., et~al.: Symmetric log-domain diffeomorphic registration: A
  demons-based approach. In: Medical Image Computing and Computer-Assisted
  Intervention -- MICCAI 2008. pp. 754--761. Springer Berlin Heidelberg,
  Berlin, Heidelberg (2008)

\bibitem{Intro_DL_Wu}
{Wu}, G., et~al.: Scalable high-performance image registration framework by
  unsupervised deep feature representations learning. IEEE Transactions on
  Biomedical Engineering  \textbf{63}(7),  1505--1516 (2016)

\bibitem{YANG2017378}
Yang, X., Kwitt, R., Styner, M., Niethammer, M.: Quicksilver: Fast predictive
  image registration – a deep learning approach. NeuroImage  \textbf{158},
  378--396 (2017)

\bibitem{zhang2021learning}
Zhang, Y., Pei, Y., Zha, H.: Learning dual transformer network for
  diffeomorphic registration. In: International Conference on Medical Image
  Computing and Computer-Assisted Intervention. pp. 129--138. Springer (2021)

\bibitem{GAN2}
Zheng, Y., et~al.: Symreg-gan: Symmetric image registration with generative
  adversarial networks. IEEE Transactions on Pattern Analysis and Machine
  Intelligence pp.~1--1 (2021)

\end{thebibliography}
\end{document}